\documentclass{llncs}
\usepackage{amsmath}
\usepackage{graphicx}
\usepackage[table,xcdraw]{xcolor}
\usepackage{amssymb}

\newcommand{\mat}[1]{\mathbf{#1}}
\DeclareMathOperator*{\argmin}{arg\,min}

\begin{document}

\title{Exploiting Partial Structural Symmetry For Patient-Specific Image Augmentation in Trauma Interventions}
%\titlerunning{Hamiltonian Mechanics} 

\author{Javad Fotouhi\inst{1,}\thanks{These authors are considered as joint first authors.}, Mathias Unberath\inst{1,*}, Giacomo Taylor\inst{1,*}, Arash Ghaani Farashahi\inst{2}, Bastian Bier\inst{1}, Russell H. Taylor\inst{3}, Greg M. Osgood, M.D.\inst{1,4}, Mehran Armand\inst{2,3,5}, and Nassir Navab\inst{1,6}}

%\authorrunning{***} % abbreviated author list (for running head)

%\tocauthor{}

\institute{
	Computer Aided Medical Procedures, Johns Hopkins University
	\and 
	Department of Mechanical Engineering, Johns Hopkins University
	\and
	Laboratory for Computational Sensing and Robotics, Johns Hopkins University
	\and
	Department of Orthopaedic Surgery, Johns Hopkins Hospital
	\and
	Applied Physics Laboratory, Johns Hopkins University
	\and
	Computer Aided Medical Procedures, Technische Universit\"{a}t M\"{u}nchen
}

\maketitle              % typeset the title of the contribution

\begin{abstract}

In unilateral pelvic fracture reductions, surgeons attempt to reconstruct the bone fragments such that bilateral symmetry in the bony anatomy is restored. We propose to exploit this "structurally symmetric" nature of the pelvic bone, and provide intra-operative image augmentation to assist the surgeon in repairing dislocated fragments. The main challenge is to automatically estimate the desired plane of symmetry within the patient's pre-operative CT. We propose to estimate this plane using a non-linear optimization strategy, by minimizing Tukey's biweight robust estimator, relying on the partial symmetry of the anatomy. Moreover, a regularization term is designed to enforce the similarity of bone density histograms on both sides of this plane, relying on the biological fact that, even if injured, the dislocated bone segments remain within the body. The experimental results demonstrate the performance of the proposed method in estimating this "plane of partial symmetry" using CT images of both healthy and injured anatomy. Examples of unilateral pelvic fractures are used to show how intra-operative X-ray images could be augmented with the forward-projections of the mirrored anatomy, acting as objective road-map for fracture reduction procedures.

\keywords{Symmetry, Robust Estimation, Orthopedics, X-ray, CT}
\end{abstract}

%%%%%%%%%%%%%%%%%%%%%%%%%%%%%%%%%%%%%%%%%%%%%%%%%%%%%%%%%%%%%%
\section{Introduction}

% good phrase: repaired pelvis, content mis-match, corrective surgery, returning symmetry, bone fragments losing alignment, self-assessment, severe trauma, missing anatomical details

% \url{https://orthoinfo.aaos.org/en/diseases--conditions/pelvic-fractures/}

% reference: karunakar2005operative
The main objective in orthopedic reduction surgery is to restore the correct alignment of the dislocated or fractured bone. In both unilateral and bilateral fractures, surgeons attempt to re-align the fractures to their natural biological alignment. In the majority of cases, there are no available anatomical imaging data prior to injury, and CT scans are only acquired after the patient is injured to identify the fracture type and plan the intervention. Therefore, no reference exists to identify the correct and natural alignment of the bone fragments. Instead, surgeons use the opposite healthy side of the patient as reference, and aim at producing symmetry across the sagittal plane~\cite{bellabarba2000distraction}. It is important to mention that, although the healthy pelvic bone is not entirely symmetric, surgeons aim at aligning the bone fragments to achieve structural and functional symmetry. For orthopedic traumatologists, the correct length, alignment, and rotation of extremities are also verified by comparing to the contralateral side. 
%Once the bone fragments are properly adjusted, the surgeons use screws, wires, and fixators to stabilize the anatomy in place. 
Examples of other fields of surgery that use symmetry for guidance include crainiofacial~\cite{vannier1984three} and breast reconstruction surgeries~\cite{edsander2001quality}.

Self-symmetry assessment is only achievable if the fractures are not bilateral, so that the contralateral side of the pelvis is intact. According to pelvis fracture classification, a large number of fracture reduction cases are only unilateral~\cite{tile1996acute}. Consequently, direct comparison of bony structures across the sagittal plane is possible in a large number of orthopedic trauma interventions.
% Iliac wing, pelvic ring, and vertical shear fractures are examples of such unilateral cases.

%Statistical modeling of anatomical structures from a population of data, based on active shape model paradigm, is introduced to register 2D truncated intra-operative C-arm X-ray images with shape atlases of large-scale CT data~\cite{sadowsky2007deformable}. 
CT-based statistical models from a population of data are particularly important when patient-specific pre-operative CT images are not present. In this situation, statistical modeling and deformable registration enable 3D understanding of the underlying anatomy using only 2D intra-operative imaging. Statistical shape models are used to extrapolate and predict the unknown anatomy in partial and incomplete medical images~\cite{chintalapani2010statistical}. In the aforementioned methods, instead of patient self-correlation, relations to a population of data are exploited for identifying missing anatomical details.
%  for patients undergoing periacetabular osteotomy

%that out of 71 feature landmarks on the pelvis, 56 of them
In this work, we hypothesize that there is a high structural correlation across the sagittal plane of the pelvis. Quantitative 3D measurements on healthy pelvis data indicate that $78.9 \%$ of the distinguishable anatomical landmarks on the pelvis are symmetric~\cite{boulay2006three}, and the asymmetry in the remaining landmarks are still tolerated for fracture reduction surgery~\cite{shen2013augmented}. To exploit the partial symmetry, we automatically detect the desired plane of symmetry using Tukey's biweight distance measure. In addition, a novel regularization term is designed that ensures a similar distribution of bone density on both sides of the plane. Regularization is important when the amount of bone dislocation is large, and Tukey's cost cannot solely drive the symmetry plane to the optimal pose. After identifying the partial symmetry, the CT volume is mirrored across the symmetry plane which then allows simulating the ideal bone fragment configurations. This information is provided intra-operatively to the surgeon, by overlaying the C-arm X-ray image with a forward-projection of the mirrored volume.  

The proposed approach relies on pre-operative CT scans of the patient. It is important to note that acquiring pre-operative CT scans is standard practice in severe trauma and fracture reduction cases. Therefore, it is valid to assume that pre-operative imaging is available for the types of fractures discussed in this manuscript, namely illiac wing, superior and inferior pubic ramus, lateral compression, and vertical shear fractures. In this paper, we introduce an approach that enables the surgeon to use patient CT scans intra-operatively, without explicitly visualizing the 3D data, but instead using 2D image augmentation on commonly used X-ray images. 
%Symmetry-based approach could also become possible for patient-specific total hip arthroplasty, in cases where pre-operative CT imaging is available~\cite{lubovsky2010acetabular}.

%%%%%%%%%%%%%%%%%%%%%%%%%%%%%%%%%%%%%%%%%%%%%%%%%%%%%%%%%%%%%%
\section{Materials and Methods}

\subsection{Problem Formulation}
%- Non-linear minimization

The plane of symmetry of an object $O \subseteq \mathbb{R}^3 $ is represented using an involutive isometric transformation $\mat{M}(g) \in E_o(3)$, such that $E_o(3) = \{ h \in E(3) \enspace \vert \enspace h (O) = O \}$, where $E(3)$ is the 3D Euclidean group, consisting of all isometries of $\mathbb{R}^3$ which map $\mathbb{R}^3$ onto $\mathbb{R}^3$. 
The transformation $\mat{M}(g)$ mirrors the object $O$ across the plane as; $\mat{o_{-x}} = \mat{M}(g) \, \mat{o_{x}}$, where $\mat{o_x} \subseteq \mathbb{P}^3$ and $\mat{o_{-x}} \subseteq \mathbb{P}^3$ are sub-volumes of object $O$ on opposite sides of the symmetry plane, and are defined in the 3D projective space $\mathbb{P}^3$.  Assuming the plane of symmetry is the Y-Z plane, $\mat{M}(g)$ is given by  $\mat{M}(g) := g \, \mat{F}_x \, g^{-1}$, where $\mat{F}_x$ is reflection about the X-axis, and $g \in SE(3)$, where $SE(3)$ is the 3D special Euclidean group. We propose to estimate $\mat{M}(g)$ by minimizing a distance function $D (\mat{M}(g))$ as:

%A function $g: \mathbb{R}^3 \rightarrow \mathbb{R}^3$ is an isometry if $||g(x)||_2 = ||x||_2$, for all $x \in \mathbb{R}^3$, when $||.||_2$ is the $l_2$-norm on $\mathbb{R}^3$. 

% for each $g \in \mathbb{SE}(3)$, $\mat{M}(g)$ is defined as:

% \begin{bmatrix}
% -1 & 0 & 0 & 0 \\
% 0 & 1 & 0 & 0 \\
% 0 & 0 & 1 & 0 \\
% 0 & 0 & 0 & 1
% \end{bmatrix}

\begin{equation}\label{eq:cost}
\argmin_{g} D(\mat{M}(g)) := d_{I}(\mat{o_{x}}, \mat{M}(g)\mat{o_{x}}) + \lambda . d_{D}(\mat{o_{x}}, \mat{M}(g)\mat{o_{x}}),
\end{equation}
combining an intensity-based distance measure $d_{I}(.)$, and a regularization term $d_{D}(.)$ based on the bone density distribution. % and is formulated by the following equation:
% \begin{equation}\label{eq:cost_prime}
% D(\mat{M}(g)) = d_{I}(\mat{o_{x}}, \mat{M}(g)\mat{o_{x}}) + \lambda . d_{D}(\mat{o_{x}}, \mat{M}(g)\mat{o_{x}}),
% \end{equation}
The term $\lambda$ in Eq.~\ref{eq:cost} is a relaxation factor, and $\lambda, d_{I}(.), d_{D}(.) \in \mathbb{R}$. Derivations of $d_{I}(.)$ and $d_{D}(.)$ are explained in Sec.~\ref{sec:tuk} and Sec.~\ref{sec:reg}, respectively. In Sec.~\ref{sec:augmentation}, we suggest an approach to incorporate the knowledge from the plane of symmetry, and provide patient-specific image augmentation in fracture reduction interventions. 

\subsection{Robust Loss for Estimation of Partial Symmetry} \label{sec:tuk}

The CT data of a pelvis only exhibits partial symmetry, as several regions within the CT volume may not have a symmetric counterpart on the contralateral side. These outlier regions occur either due to dislocation of the bone fragments, or asymmetry in the natural anatomy. To estimate the plane of "partial symmetry", we suggest to minimize a disparity function that is robust with respect to outliers, and only considers the partial symmetry present in the volumetric data. The robustness to outlier is achieved by down-weighting the error measurements associated to potential outlier regions. To this end, we estimate the plane of partial symmetry by minimizing Tukey's beweight loss function defined as~\cite{huber2011robust}:
%mosteller1977data

\begin{equation}
\begin{aligned}
&d_{I}(\mat{o_{x}}, \mat{M}(g) \mat{o_{x}}) = \sum_{i=1}^{ |\Omega_s|} \frac{\rho (e_i(\mat{M}(g)))}{|\Omega_s|},\\
&\rho (e_i(\mat{M}(g))) = \begin{cases}
	e_i(\mat{M}(g))\left[ 1 - \left( \frac{e_i(\mat{M}(g))}{c} \right)^2 \right]^2 &;  |e_i(\mat{M}(g))| \leqslant c,\\
    0 &;  \text{otherwise},
\end{cases}
\end{aligned}
\end{equation}
% that map using the transformation $\mat{M}(g)$ to voxel elements inside the CT volume and on the opposite side of symmetry plane
with $\Omega_s$ being the spatial domain of CT elements. The threshold of assigning data elements as outlier is defined by a constant factor $c$ that is inversely proportional to the down-weight assigned to outliers. As suggested in the literature, $c = 4.685$ provides high asymptotic efficiency~\cite{huber2011robust}. The residual error for the $i$-th voxel element is $\rho (e_i(\mat{M}(g)))$ and is defined as following:

\begin{equation}\label{e(M)}
\begin{aligned}
&e_i(\mat{M}(g)) = \frac{| \text{CT}(o_{x_i}) - \text{CT}(\mat{M}(g) o_{x_i}) |}{S},\\
& S = \frac{\text{Median}  \left\{ e_i(\mat{M}(g)) \right\}_{i=1}^{|\Omega_s|} }{0.6745}.
\end{aligned}
\end{equation}
In Eq.~\ref{e(M)}, $o_{x_i}$ is the $i$-th voxel element, $\text{CT}(o_{x_i})$ is the intensity of $o_{x_i}$, $S$ is the scaling factor corresponding to the standard deviation of the residual error, and 0.6745 represents the one-half of the interquantile range of a normal distribution. 

% Robust loss
% A paper about Robust estimator~\cite{kumar1994robust}

\subsection{Bone Density Histogram Regularization}\label{sec:reg}

\begin{figure}
  \centering
  \includegraphics[width=\textwidth]{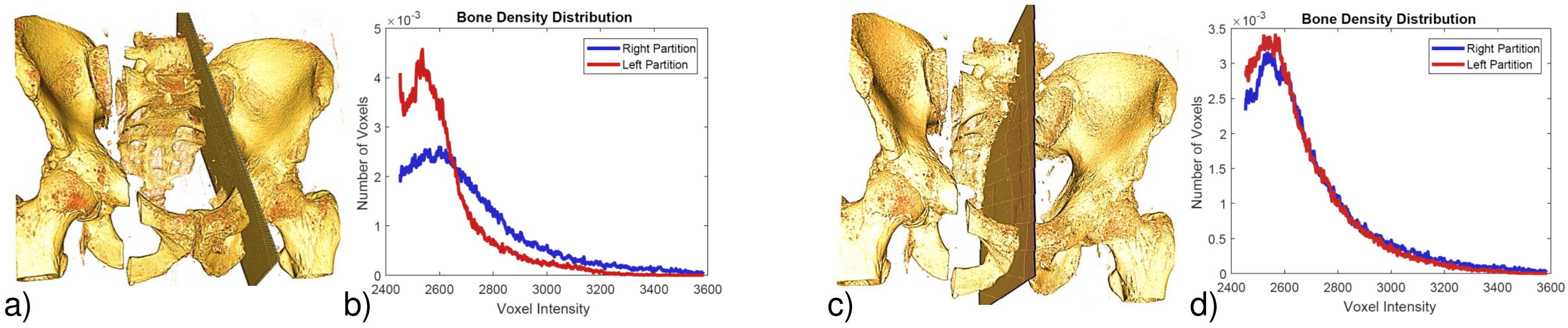}
  \caption{Bone intensity histograms are shown in \textbf{(a-b)} and {\textbf{(c-d)}} before and after estimating the plane of partial symmetry.}
    \label{fig:histogram}
\end{figure}

In fractured bones, the distribution of bone material inside the body will remain nearly unaffected. Based on this fact, we hypothesize that the distribution of bone intensities, \textit{i.e.} histograms of bone Hounsfield Unit (HU), on the opposite sides of the plane of symmetry remains similar in presence of fracture (example shown in Fig.~\ref{fig:histogram}). Therefore, we design a regularization term based on normalized mutual information as follows~\cite{studholme1999overlap}:
\begin{equation}\label{NMI}
d_{D}(\mat{o_{x}}, \mat{M}(g) \mat{o_{x}}) = - \frac{H(\text{CT}(\mat{o_{x}})) + H(\text{CT}(\mat{M}(g) \mat{o_{x}}))}{H(\text{CT}(\mat{o_{x}}), \text{CT}(\mat{M}(g) \mat{o_{x}})))},
\end{equation}
where $H(.)$ is the entropy of voxels' HU distribution. Minimizing the distance function in Eq.~\ref{NMI} is equivalent to increasing the similarity between the distributions of bone on the opposing sides of the plane of partial symmetry. 

\subsection{Patient-Specific Image Augmentation}\label{sec:augmentation}
\begin{figure}
  \centering
  \includegraphics[width=\textwidth]{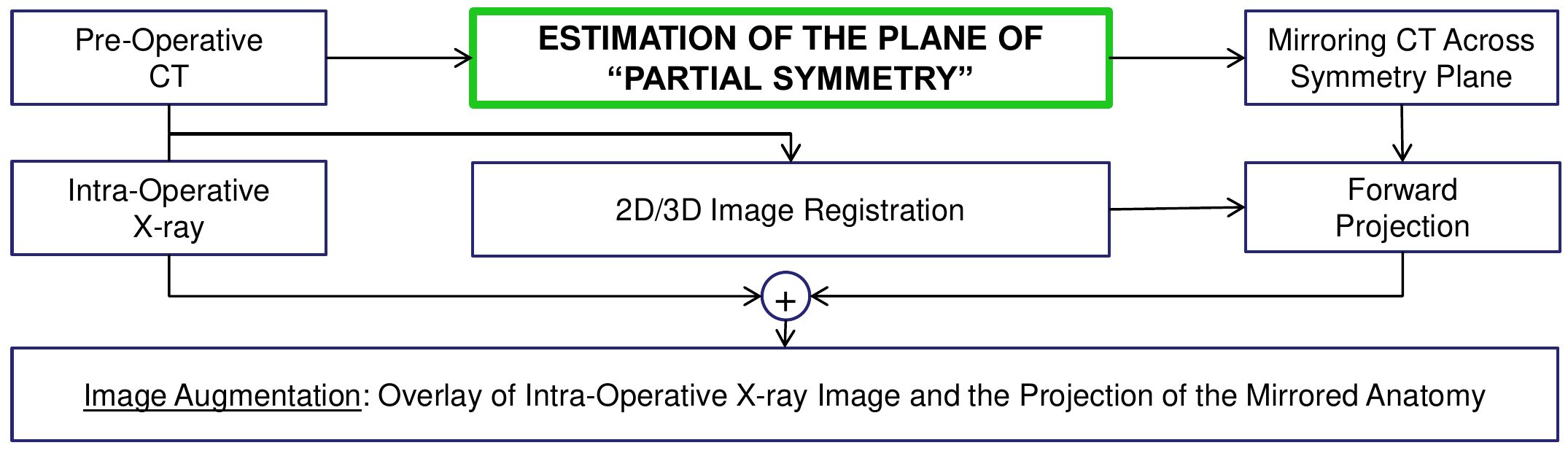}
  \caption{Workflow for patient-specific image augmentation based on partial symmetry}
    \label{fig:workflow}
\end{figure}

After estimating the plane of partial symmetry, the CT volume is mirrored across this plane to construct a patient-specific CT volume representing the bony structures "as if they were repaired". It is important to note that, although the human pelvic skeleton is not entirely symmetric, it is common in trauma interventions to consider it as symmetric, and use the contralateral side as reference. 

To assist the orthopedic surgeon in re-aligning bone fragments, we propose to augment intra-operative X-ray images with the bone contours from the mirrored CT volume. This step requires generation of digitally reconstructed radiographs (DRRs) from views identical to the one acquired intra-operatively using a C-arm. Hence, 2D/3D image registration based on normalized cross-correlation (NCC) is employed to estimate the projection geometry between the intra-operative X-ray image and the pre-operative patient CT data. This projective transformation is then used to forward-project and generate DRR images from the same viewing angle that the X-ray image was acquired. Finally, we augment the X-ray image with the edge-map acquired from the DRR that will then serve as a road-map for re-aligning the fragmented bones. The proposed workflow is shown in Fig.~\ref{fig:workflow}.

% - Gradient weighted forward projection

%- When discussing 2D/3D registration, address the initialization problem and that it could be considered as highly ill-posed. but recent work shows that vision-based inforamtion that provide tracking of the surgical scene could increase the success rate for 2D-3D registration significantly~\cite{fotouhi2017pose}. \\

%- Is it possible to adaptively change the value of sigma over iterations (similar to changing threshold value in ICP)\\

%%%%%%%%%%%%%%%%%%%%%%%%%%%%%%%%%%%%%%%%%%%%%%%%%%%%%%%%%%%%%%
\section{Experimental Validation and Results}\label{sec:results}

%\todo{Figure of bone histogram, before and after alignment}\\
%\todo{Figure of the synthetic data with different amount of Gaussian noise, and different amount of outlier} 
%\todo{How are the bones separated in the histogram}

We conducted experiments on synthetic and real CT images of healthy and fractured data. 
%In the following section we provide the results for each of these experiments. Finally, we demonstrate the patient-specific image augmentation based on the partial symmetry. 
For all experiments, the optimization was performed using bound constrained by quadratic approximation algorithm, where the maximum number of iterations was set to $100$. The regularization term $\lambda$ in Eq.~\ref{eq:cost} was set to $0.5$ which allowed $d_I(.)$ to be the dominant term driving the similarity cost, and $d_{D}(.)$ to serve as a data fidelity term.
% ~\cite{powell2009bobyqa}

\begin{figure}
  \centering
  \includegraphics[width=\textwidth]{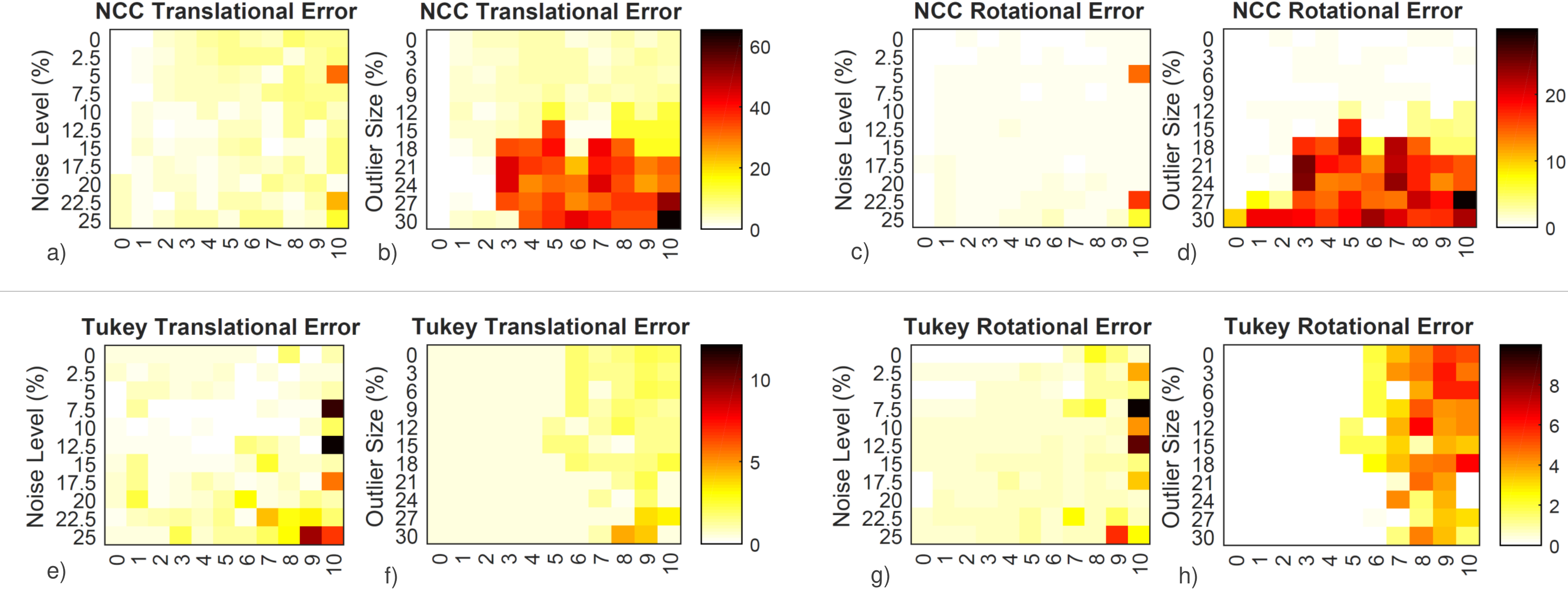}
  \caption{The performance of Tukey-based distance measure is evaluated against noise level and amount of outlier. The horizontal axis in each plot indicates the distance of the plane at the initial step to the ground-truth symmetry plane. The initialization parameters were increased with the increments of $2$ voxels translation and $2^{\circ}$ rotation along each axis. The amount of outlier was varied between $0\%$ to $30\%$ of the size of the entire volume, and the Gaussian noise between $0 \%$ to $25\%$ of the highest intensity in the volume. \textbf{(a-d)} and \textbf{(e-h)} show the performance of NCC and Tukey robust estimator, respectively. It is important to note that different heat-map color scales are used for NCC and Tukey to demonstrate the changes within each sub-plot.}
    \label{fig:sensitivity}
\end{figure}

\paragraph{Performance Evaluation on Synthetic Data:}
A synthetic 3D data of size $100^3$ voxels and known plane of symmetry was generated. We evaluated the performance of the Tukey-based cost $d_I(.)$ with respect to noise and outliers, and compared the outcome to NCC-based cost. The results are shown in Fig.~\ref{fig:sensitivity}.
% with perfect symmetry along the Y-Z plane

\paragraph{Estimating the Plane of Partial Symmetry on Healthy Data:}
The plane of partial symmetry was estimated for twelve CT datasets with no signs of fractures or damaged bone. Four of the volumes were lower torso cadaver CT data, and eight were from subjects with Sarcoma. After estimating the plane of partial symmetry, the CT volumes were mirrored across the estimated plane. We then identified the following 4 anatomical landmarks and measured the distance between each landmark on the original volume to the corresponding landmark on the mirrored CT volume: \textbf{L$_1$}: anterior superior iliac spine, \textbf{L$_2$}: posterior superior iliac spine, \textbf{L$_3$}: ischial spine, and \textbf{L$_4$}: ischial ramus. Results of this experiment using NCC, Tukey robust estimator $d_I(.)$, and regularized Tukey $d_I(.) + \lambda d_D(.)$ distance functions are shown in Table.~\ref{table:healthy}.   

\begin{table}[t]
\centering
\caption{Errors in estimation of partial symmetry were measured using four anatomical landmarks. The values in the table are in mm units, and are shown as mean $\pm$ SD. The final row are the results of our proposed method.}
\label{table:healthy}
\begin{scriptsize}
\begin{tabular}{|l|c|c|c|c|}
\cline{2-5}
\multicolumn{1}{c|}{} & \textbf{L$_1$} & \textbf{L$_2$} & \textbf{L$_3$} & \textbf{L$_4$} \\ \hline
\textbf{NCC} & $12.8 \pm 7.15$ & $11.5 \pm 10.7$ & $7.81 \pm 5.08$ & $7.12 \pm 5.20$ \\
\textbf{Tukey} & $6.79 \pm 3.95$ & $6.95 \pm 5.34$ & $4.63 \pm 2.60$ & $4.75 \pm 1.85$ \\
\textbf{Regularized Tukey} & $\mathbf{3.85 \pm 1.79}$ & $\mathbf{4.06 \pm 3.32}$ & $ \mathbf{3.16 \pm 1.41}$ & $\mathbf{2.77 \pm 2.13}$ \\ \hline
\end{tabular}
\end{scriptsize}
\end{table}

\begin{figure}[t]
  \centering
  \includegraphics[width=0.98\textwidth]{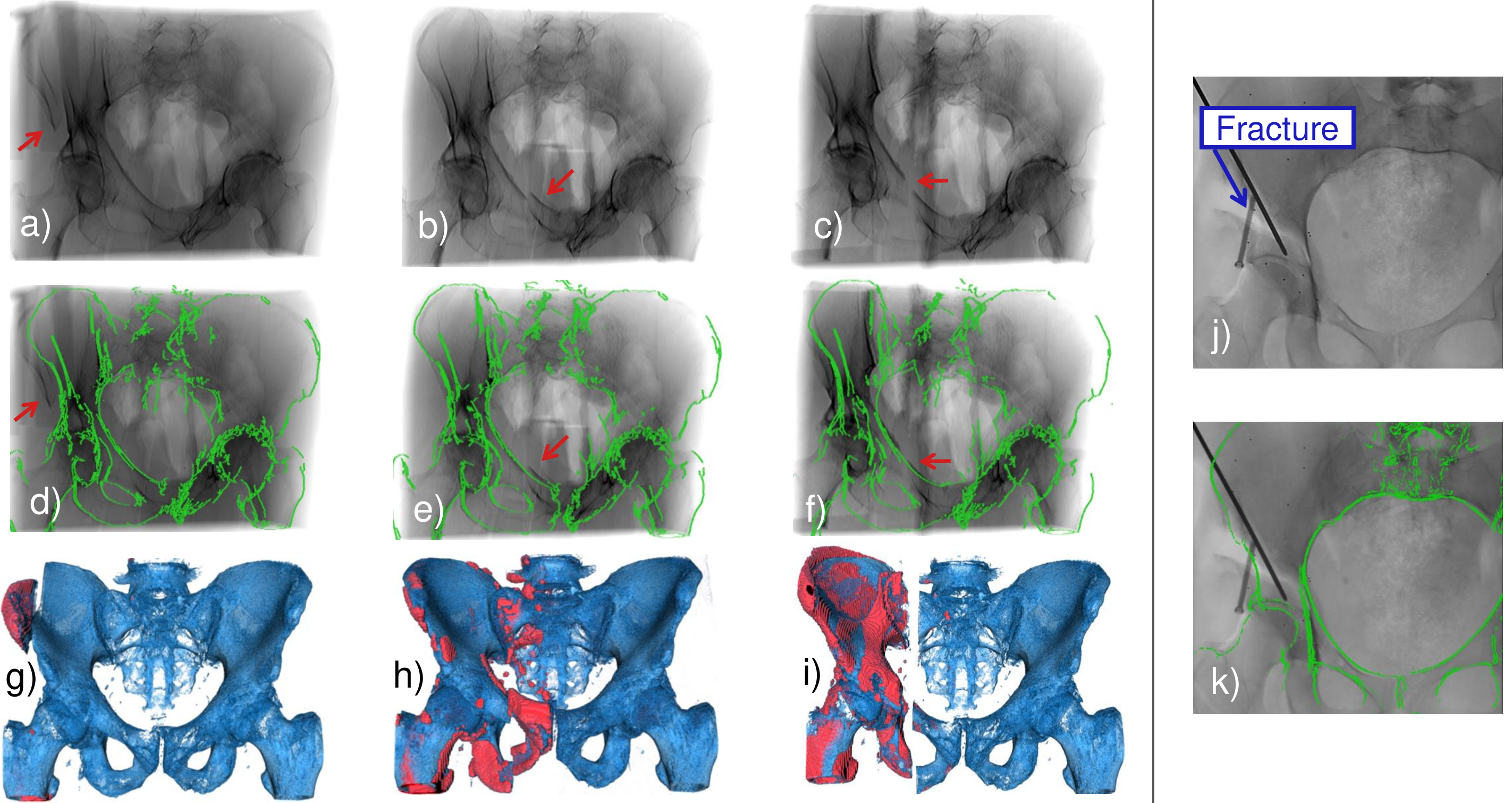}
  \caption{\textbf{(a-c)} are simulated iliac wing, pelvic ring, and vertical shear fractures, respectively. Image augmentations are shown in \textbf{(d-f)}. The red arrows indicate the fracture location in the pelvis. The green contours represent the desired bone contours to achieve bilateral symmetry. \textbf{(g-h)} show the 3D visualization of the fractures, where red color indicates regions that were rejected as outliers using Tukey's robust estimator. The intra-operative X-ray image in \textbf{(j)} is augmented with the edge-map of the DRR generated from the mirrored volume using the projection geometry estimated from 2D/3D X-ray to CT image registration\textbf{(k)}.}
    \label{fig:augmented}
\end{figure}

\paragraph{Estimating the Plane of Partial Symmetry on Fractured Data:}
We simulated three different fractures - \textit{i.e.} iliac wing, pelvic ring, and vertical shear fractures (shown in Fig.~\ref{fig:augmented}(a-c)) - and evaluated the performance of the proposed solution in presence of bone dislocation. These three fractures were applied to three different volumes, and in total nine fractured CT volumes were generated. The error measurements are presented in Table~\ref{table:fractured}.  
% Error measurements on fractured data reported as mean $\pm$ SD in mm units

\begin{table*}
\centering
\caption{Error measurements on fractured data in mm units.}
\label{table:fractured}
\begin{scriptsize}
\begin{tabular}{l|c|c|c|c|}
\cline{2-5}
\multicolumn{1}{c|}{\textbf{}} & \textbf{L$_1$} & \textbf{L$_2$} & \textbf{L$_3$} & \textbf{L$_4$} \\ \hline
\multicolumn{5}{|c|}{\cellcolor[HTML]{ECF4FF}\textbf{Iliac Wing Fracture}} \\
\multicolumn{1}{|l|}{\textbf{NCC}} & $26.4 \pm 14.3$ & $20.1 \pm 14.2$ & $14.9 \pm 9.17$ & $10.2 \pm 1.04$ \\
\multicolumn{1}{|l|}{\textbf{Tukey}} & $6.36 \pm 3.40$ & $6.80 \pm 4.10$ & $6.90 \pm 5.86$ & $4.89 \pm 1.29$ \\
\multicolumn{1}{|l|}{\textbf{Regularized Tukey}} & $\mathbf{3.60 \pm 2.93}$ & $\mathbf{3.30 \pm 3.13}$ & $\mathbf{4.01 \pm 1.24}$ & $\mathbf{2.06 \pm 0.92}$ \\ \hline
\multicolumn{5}{|c|}{\cellcolor[HTML]{ECF4FF}\textbf{Pelvic Ring Fracture}} \\
\multicolumn{1}{|l|}{\textbf{NCC}} & $39.2 \pm 39.9$ & $27.0 \pm 23.9$ & $26.6 \pm 32.5$ & $25.4 \pm 31.0$ \\
\multicolumn{1}{|l|}{\textbf{Tukey}} & $4.54 \pm 2.39$ & $6.14 \pm 5.88$ & $6.28 \pm 3.81$ & $3.68 \pm 0.81$ \\
\multicolumn{1}{|l|}{\textbf{Regularized Tukey}} & $\mathbf{2.17 \pm 1.37}$ & $\mathbf{3.75 \pm 3.17}$ & $\mathbf{2.03 \pm 0.73}$ & $\mathbf{1.98 \pm 0.99}$ \\ \hline
\multicolumn{5}{|c|}{\cellcolor[HTML]{ECF4FF}\textbf{Vertical Shear Fracture}} \\
\multicolumn{1}{|l|}{\textbf{NCC}} & $28.1 \pm 15.1$ & $16.6 \pm 11.9$ & $21.7 \pm 6.91$ & $19.7 \pm 6.02$ \\
\multicolumn{1}{|l|}{\textbf{Tukey}} & $15.8 \pm 7.66$ & $10.9 \pm 8.48$ & $11.1 \pm 3.54$ & $11.4 \pm 5.52$ \\
\multicolumn{1}{|l|}{\textbf{Regularized Tukey}} & $\mathbf{5.46 \pm 2.04}$ & $\mathbf{3.52 \pm 2.63}$ & $\mathbf{4.85 \pm 1.86}$ & $\mathbf{5.28 \pm 2.89}$ \\ \hline
\end{tabular}
\end{scriptsize}
\end{table*}

\paragraph{Intra-Operative Image Augmentation:} 
After estimating the plane of partial symmetry, we mirrored the healthy side of the pelvis across the plane of partial symmetry. In Fig.~\ref{fig:augmented}(d-f) we present the superimposition of the fractured data shown in Fig.~\ref{fig:augmented}(a-c) with the edge-map extracted from gradient-weighted DRRs. Moreover, we preformed 2D/3D intensity-based registration between the pre-operative CT and the intra-operative X-ray (Fig.~\ref{fig:augmented}j), and used the estimated projection geometry to generate DRRs and augment the X-ray image of the fractured bone. The augmentation is shown in Fig.~\ref{fig:augmented}k. 
% This augmentation can assist the surgeon in re-aligning bone fragments.

%%%%%%%%%%%%%%%%%%%%%%%%%%%%%%%%%%%%%%%%%%%%%%%%%%%%%%%%%%%%%%
\section{Discussion and Conclusion}
This work presents a novel method to estimate partial structural symmetry in CT images of fractured pelvises. We used Tukey's biweight robust estimator which prevents outlier voxel elements from having large effects on the similarity measure. Moreover, Tukey's distance function is regularized by enforcing high similarity in bilateral bone HU distribution. The experimental results on synthetic data indicate that Tukey-based similarity cost outperformed NCC-based similarity cost substantially in the presence of noise and outliers. The results in Table~\ref{table:healthy} show an average landmark error of $5.78$\,mm and $3.46$\,mm using Tukey- and regularized Tukey-based cost, respectively. Similarly for fractured data presented in Table~\ref{table:fractured}, the mean error reduced from $7.91$\,mm to $3.50$\,mm after including the regularization term.  

%The results in Tables~\ref{table:healthy} and~\ref{table:fractured} indicate low target registration errors for both healthy and fractured data using the proposed approach. 

In conclusion, we proposed to incorporate the knowledge from partial symmetry and provide intra-operative image augmentation to assist orthopedic surgeons in re-aligning the bone fragments with respect to bilateral symmetry. Our work relies on pre-operative CT images and is only applicable to surgical interventions where pre-operative 3D imaging exist. This solution enables patient-specific image augmentation which is not possible using statistical atlases. Using atlases for this application requires a large population of patient pelvis data for different age, sex, race, disease, etc. which are not available. The suggested patient-specific image augmentation could be applicable in reducing pelvic fractures, particularly when dislocations are large and external fixators are used to stabilize the bone.


\begin{thebibliography}{1}
	
	\bibitem{bellabarba2000distraction}
	Bellabarba, C., Ricci, W.M., Bolhofner, B.R.:
	\newblock Distraction external fixation in lateral compression pelvic
	fractures.
	\newblock Journal of orthopaedic trauma \textbf{14}(7) (2000)  475--482
	
	\bibitem{vannier1984three}
	Vannier, M.W., Marsh, J.L., Warren, J.O.:
	\newblock Three dimensional ct reconstruction images for craniofacial surgical
	planning and evaluation.
	\newblock Radiology \textbf{150}(1) (1984)  179--184
	
	\bibitem{edsander2001quality}
	Edsander-Nord, A., Brandberg, Y., Wickman, M.:
	\newblock Quality of life, patients' satisfaction, and aesthetic outcome after
	pedicled or free tram flap breast surgery.
	\newblock Plastic and reconstructive surgery \textbf{107}(5) (2001)  1142--53
	
	\bibitem{tile1996acute}
	Tile, M.:
	\newblock Acute pelvic fractures: I. causation and classification.
	\newblock JAAOS-Journal of the American Academy of Orthopaedic Surgeons
	\textbf{4}(3) (1996)  143--151
	
	\bibitem{chintalapani2010statistical}
	Chintalapani, G., Murphy, R., Armiger, R.S., Lepisto, J., Otake, Y., Sugano,
	N., Taylor, R.H., Armand, M.:
	\newblock Statistical atlas based extrapolation of ct data.
	\newblock In: Medical Imaging 2010: Visualization, Image-Guided Procedures, and
	Modeling. Volume 7625., International Society for Optics and Photonics (2010)
	762539
	
	\bibitem{boulay2006three}
	Boulay, C., Tardieu, C., B{\'e}naim, C., Hecquet, J., Marty, C., Prat-Pradal,
	D., Legaye, J., Duval-Beaup{\`e}re, G., P{\'e}lissier, J.:
	\newblock Three-dimensional study of pelvic asymmetry on anatomical specimens
	and its clinical perspectives.
	\newblock Journal of Anatomy \textbf{208}(1) (2006)  21--33
	
	\bibitem{shen2013augmented}
	Shen, F., Chen, B., Guo, Q., Qi, Y., Shen, Y.:
	\newblock Augmented reality patient-specific reconstruction plate design for
	pelvic and acetabular fracture surgery.
	\newblock International journal of computer assisted radiology and surgery
	\textbf{8}(2) (2013)  169--179
	
	\bibitem{huber2011robust}
	Huber, P.J.:
	\newblock Robust statistics.
	\newblock In: International Encyclopedia of Statistical Science.
	\newblock Springer (2011)  1248--1251
	
	\bibitem{studholme1999overlap}
	Studholme, C., Hill, D.L., Hawkes, D.J.:
	\newblock An overlap invariant entropy measure of 3d medical image alignment.
	\newblock Pattern recognition \textbf{32}(1) (1999)  71--86
	
\end{thebibliography}
\end{document}